\documentclass{article}

\usepackage{arxiv}

\usepackage[utf8]{inputenc} % allow utf-8 input
\usepackage[T1]{fontenc}    % use 8-bit T1 fonts
\usepackage{hyperref}       % hyperlinks
\usepackage{url}            % simple URL typesetting
\usepackage{booktabs}       % professional-quality tables
\usepackage{amsfonts}       % blackboard math symbols
\usepackage{nicefrac}       % compact symbols for 1/2, etc.
\usepackage{microtype}      % microtypography
\usepackage{graphicx}
\usepackage{natbib}
\usepackage{xcolor}
\usepackage{tikz}
\usepackage{algorithm}
\usepackage{algpseudocode}
\input{generic_aliases}

\newcommand{\alex}[1]{{\color{blue} #1}}

\title{Diagnosis Uncertain Models For Medical Risk Prediction}

\author{Alexander Peysakhovich \\
\texttt{alex.peys@gmail.com}
	 \And
	Rich Caruana \\
	\texttt{Microsoft Research}
	\And
	Yin Aphinyanaphongs \\
	\texttt{NYU Langone}
}

\renewcommand{\shorttitle}{\textit{arXiv} Template}

\begin{document}
\maketitle

\begin{abstract}
We consider a patient risk models which has access to patient features such as vital signs, lab values, and prior history but does not have access to a patient's diagnosis. For example, this occurs in a model deployed at intake time for triage purposes. We show that such `all-cause' risk models have good generalization across diagnoses but have a predictable failure mode. When the same lab/vital/history profiles can result from diagnoses with different risk profiles (e.g. E.coli vs. MRSA) the risk estimate is a probability weighted average of these two profiles. This leads to an under-estimation of risk for rare but highly risky diagnoses. We propose a fix for this problem by explicitly modeling the uncertainty in risk prediction coming from uncertainty in patient diagnoses. This gives practitioners an interpretable way to understand patient risk beyond a single risk number.
\end{abstract}

\section{Introduction}
In-hospital patient outcome prediction is a major research area at the intersection of machine learning and medicine \citep{barfod2012abnormal,taylor2016prediction,brajer2020prospective,naemi2021machine,soffer2021predicting,wiesenfeld2022ai}. An important application of such models is `early' risk prediction - for example, using risk scores for triage \citep{raita2019emergency, klug2020gradient}. Early prediction often requires calculating patient risk when primary diagnosis is still unknown or uncertain. We propose a method for incorporating uncertainty about diagnosis into mortality risk assessments in an interpretable and actionable way.

We study the problem of all-cause in-hospital mortality prediction in the MIMIC-IV dataset \citep{johnson2023mimic}. We find that a single model which pools all data and ignores diagnoses (we refer to this as the all-cause model or ACM) performs better at prediction than diagnosis-specific modeling. This increase in performance comes from the fact that the ACM has access to more data (so has lower variance) and that there is substantial transferrability in risk across diagnoses (so the ACM bias is not that high). We see this even more starkly by showing that a model trained only on out-of-diagnosis data can, due to this logic, predict risk within a diagnosis just as well as a model trained on that diagnosis only.

While ACM are on average quite performant, we find that there are cases where they can fail. The law of total probability tells us we can factorize the probability of mortality given features $x$ (but not diagnosis $d$) as an average of the diagnosis-specific probabilities weighted by the probability of diagnoses conditional on that feature value $p(y \mid x) = \sum_{d} p(y \mid x, d) p(d \mid x)$. That is, an all-cause risk model is an average of the risk that would have been predicted for the lab/vital/history measurements given a cause, weighted by the probability of each cause.

From here we see the main issue - when multiple diagnoses are consistent with a lab/vital/history profile $x$ the ACM prediction is an average of them. In a case where some of the diagnoses are relatively benign but one or more is risky, the averaging will `de-risk' the truly risky patients. We show that this occurs in the MIMIC-IV dataset.

It is typical in medical practice to consider treatment based on `risk weighted' probabilities - consider the case of a patient who presents with a tick bite. Though the posterior probability of lyme disease is relatively low, many practitioners would still treat the patient with a prophylactic antibiotic\footnote{See \href{https://www.cdc.gov/ticks/tickbornediseases/tick-bite-prophylaxis.html}{https://www.cdc.gov/ticks/tickbornediseases/tick-bite-prophylaxis.html} for the CDC guidance on this topic.} as the treatment is safe and the risk to the patient from doing nothing is high in the `worst' case where transmission has indeed occurred.

Our main technical contribution is to take a similar logic to risk modeling. We exploit the fact that our problem is such that \textit{at training time} we do have access to a patient's diagnosis, even though we do not have access to this diagnosis \textit{at inference time}. Therefore, we propose to explicitly incorporate uncertainty about the diagnosis into our predictions.

\begin{figure}[h!]
    \centering
    \includegraphics[scale=.5]{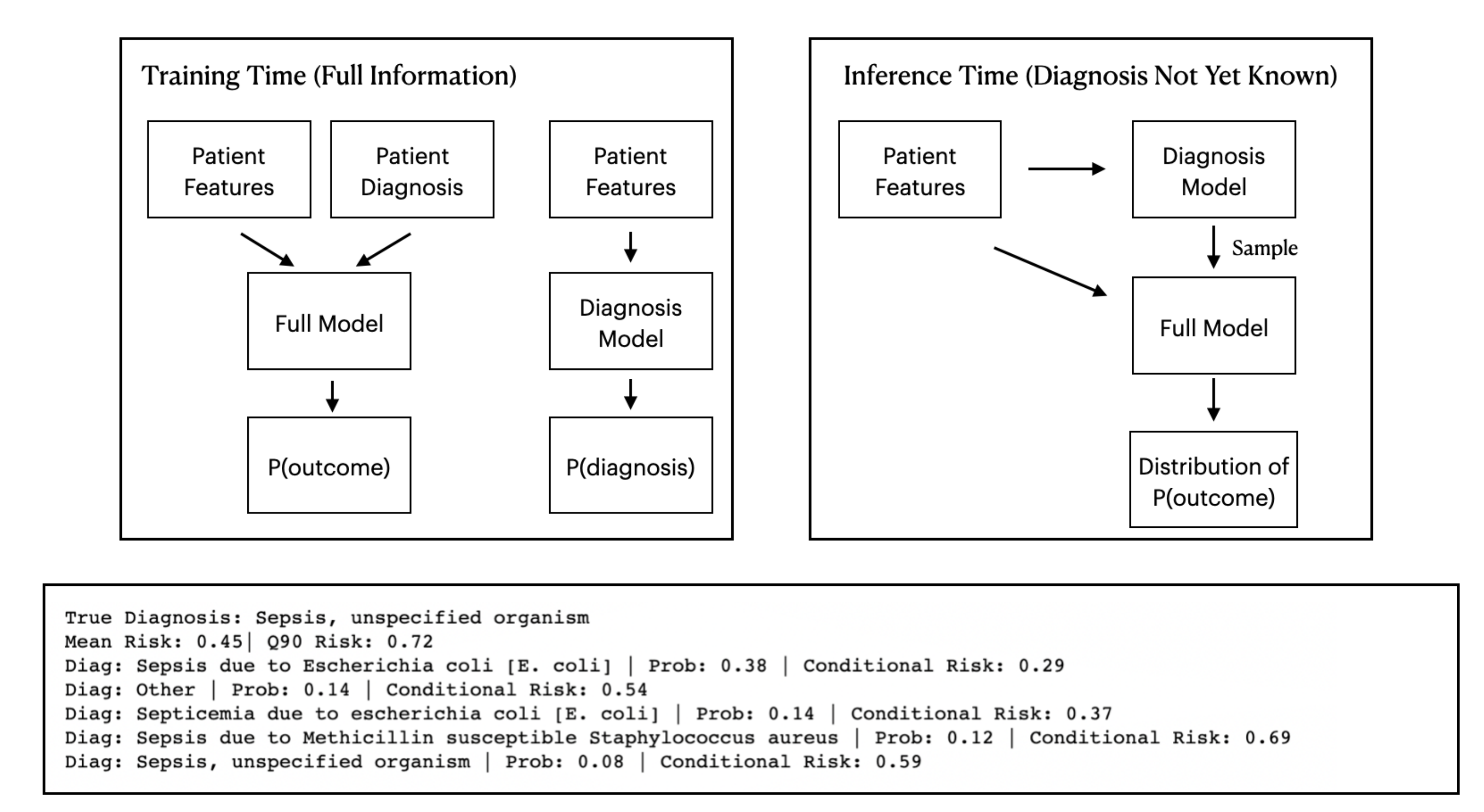}
    \caption{Panel 1: DU-ACM architecture. Panel 2: DU-ACM example of a patient where the risk model estimates change substantially due to uncertainty in underlying diagnosis.}
    \label{fig:du_acm_architecture}
\end{figure}

At test time we train two models. One that predicts risk using features and diagnosis. The second that predicts diagnosis using features. At inference time (when diagnosis is unknown) we construct a distribution of possible patient diagnoses by sampling from the diagnosis model and getting a risk prediction for each of them combined with the features (see Figure \ref{fig:du_acm_architecture}. We refer to this as a diagnosis-uncertain all-cause model or DU-ACM.

We show that the DU-ACM allows for interpretable outputs: we can output both the patient's expected risk as well as a pessimistic (e.g. $90^{th}$ percentile) estimate. Importantly, we can also surface to the user which diagnoses are leading to the pessimistic insight. Panel 2 in Figure~\ref{fig:du_acm_architecture} shows an example of a patient from the MIMIC-IV dataset where the DU model is highly uncertain of their risk status - the diagnosis model (correctly) identifies the patient has an infection however from the vitals/labs available to it it cannot tell whether the infection is MRSA (antibiotic resistant) or E.coli (relatively easily treatable). This makes the risk model very uncertain - if the underlying diagnosis is, in fact, MRSA then the patient is extremely high risk. If it is E.coli, however, the risk is much lower.

The way that the DU-ACM is structured allows practitioners to easily interact with the model and rule out (or confirm) certain diagnoses. We can then sample from the new implied diagnosis distribution to get an updated DU-ACM risk prediction. The example above suggests that ruling out (or confirming) MRSA is an important step in figuring out whether the patient is high or low risk.

In the rest of the paper we discuss the theory of the DU-ACM, try various architectures, and show examples on the MIMIC-IV dataset. We note that our main contribution is agnostic to the model architecture used to fit it (e.g. logistic regression vs. neural network vs. random forest), however, we will still study this question here as it provides a useful view on the structure of our datasets and our problem.

\section{Related Work}
There is a large and growing literature at the intersection of machine learning on predicting patient in-hospital risk \citep{barfod2012abnormal,taylor2016prediction,brajer2020prospective,naemi2021machine,soffer2021predicting,wiesenfeld2022ai}, deterioration \citep{brekke2019value,escobar2020automated,gerry2020early}, and post-discharge re-admission \citep{jamei2017predicting,mahmoudi2020use}. A large fraction of this literature focuses on feature selection \cite{brekke2019value,barda2020developing}, model architecture \citep{christodoulou2019systematic} and model portability \citep{wiesenfeld2022ai}. While we touch on these topics in our analyses, our main focus is the increasing the interpretability and usability of risk models by incorporating uncertainty estimates driven by uncertainty about diagnosis. The DU approach is a complement, not a substitute, to existing approaches.

A large fraction of the literature focuses on diagnosis specific risk models \citep{taylor2016prediction,edwards2016development,barda2020developing}. There are several reasons for this preference including because such models allow for the use of diagnosis-specific features which allows for better model performance. Our results on ACM vs. diagnosis-specific modeling suggest that there are gains to be had from training multiple such models together in a multi-task fashion \citep{caruana1997multitask,harutyunyan2019multitask} as there is substantial information pooling. In addition, the DU idea can be applied to more complex models when diagnosis is uncertain (sampling from some predictive model of the diagnosis) or some features are unavailable (sampling from an imputation model and considering `pessimistic' outcomes). Given the success of large `foundation models' \citep{bommasani2021opportunities} in other areas of AI, such multi-task risk models appear to be an interesting area for future research.

Uncertainty in model predictions has been a major topic of study across statistics, machine learning, and many applications \citep{krzywinski2013importance,ghahramani2015probabilistic}. Much of the work divides uncertainty in `aleatoric' (due to inherent randomness) and  `epistemic' (due to lack of knowledge about some important parameter) \cite{hullermeier2021aleatoric}. The DU-ACM is a way to expose a particular type of uncertainty, which does not neatly fit into either category, and that does not go away with more data - the uncertainty which comes from diagnosis information which we will come to learn later.

There is a huge literature on dealing with missing data in statistical modeling \citep{allison2001missing,little2019statistical}. The typical setup of the missing data literature is a feature vector $x$ and a target $y$ where some elements of $x$ for some data points are missing. Missing data methods typically try to impute these missing values such that we can train a model to predict $y$ from $x$. A major question in this literature is what types of imputation methods allow for `good' (often unbiased) results under what types of assumptions \citep{little2019statistical}. The DU-ACM can be thought of as an imputation method for missing diagnoses, however, our problem has much more structure than the typical missing data problem and our goal is interpretable uncertainty quantification which is not the typical goal in many missing data papers. The logic of the DU-ACM, however, can likely be exported to many other scenarios and looking at other missing data problems is an interesting direction for future work.

Understanding `why' a machine learning model has made a particular prediction is an important question in high stakes fields such as medicine or law \citep{ahmad2018interpretable,rudin2019stop}. Interpretable machine learning is the field which studies either models which can be easily understood (sometimes called glassbox models \citet{nori2019interpretml,agarwal2021neural}) or ways to make complex (black box \citet{ribeiro2016model,lundberg2017unified}) models like neural-networks more understandable \citep{molnar2020interpretable}. The DU approach is agnostic to the type of algorithm used to construct the outcome/diagnosis predictors so can be thought of as a black-box interpretable method for a particular problem. Seeing whether DU-like logic can be applied in other black-box interpretable algorithms is an interesting question for future research.

Algorithmic bias is an important subject of study in modern machine learning \citep{barocas-hardt-narayanan}. The problem the DU approach tries to solve - the fact that in some cases high risk patients are `lumped in' with low risk ones due to a lack of knowledge about diagnosis  in the ACM - can be framed as a particular type of algorithmic bias which arises from missing data about the patient's `class membership'.

Finally, there is a large debate on `generalizability' in medical machine learning \citep{altman2000we,futoma2020myth}. Our paper contributes to this discussion by showing that, at least in MIMIC data, there is substantial across-diagnosis generalizability in an all-cause predictive model of risk but that we need to take proper consideration of failure points (in our case, diagnosis uncertainty and base-rate risk). Thus, it is insufficient to simply train a model on a large dataset, look at an improvement in some error metric, and declare victory. Understanding whether modeling explicit generalization uncertainty can help in other domains is an interesting area for future work.

\section{Theory}
We consider the problem of training a risk (e.g. mortality) prediction model on incoming patients using information readily available at intake (labs, vital signs, patient comborbidities, records of past visits, etc...). 

We call such a  model which does not explicitly include the cause of the patient's visit an \textbf{all cause model} (ACM). Such a model is useful in many cases in practice, especially for early decision-making such as triage when basic information is available but a full diagnosis has not yet been made.

\begin{figure}[h!]
\centering

\begin{tikzpicture}[>=latex]

% Define styles for the circles
\tikzstyle{state}=[circle, draw, minimum size=1cm, text width=2cm, align=center, font=\fontsize{10}{16}\selectfont,dashed]
\tikzstyle{diagnosis}=[circle, draw, minimum size=1cm, text width=2cm, align=center, font=\fontsize{10}{16}\selectfont,dashed]
\tikzstyle{vitals}=[circle, draw, minimum size=1cm, text width=2cm, align=center, font=\fontsize{8}{16}\selectfont]
\tikzstyle{outcome}=[circle, draw, minimum size=1cm, text width=2cm, align=center, font=\fontsize{10}{16}\selectfont]

% Draw the circles
\node[state] (state) at (-3.5,0) {Patient State};
\node[diagnosis] (diagnosis) at (0,0) {Diagnosis};
\node[vitals] (vitals) at (-3.5,-3.5) {Vitals/Labs/Patient Records};
\node[outcome] (outcome) at (0,-3.5) {Outcome};

% Draw the arrows
\draw[->] (state) -- (vitals);
\draw[->] (state) -- (outcome);
\draw[->] (state) -- (diagnosis);
\draw[->] (vitals) -- (diagnosis);
\draw[->] (diagnosis) -- (outcome);

\end{tikzpicture}

\caption{A simplified model of our generating process. The ACM, which is used early on in the stay, is only able to observe a patient's vitals/labs/records and does not observe the full latent state or the diagnosis (which is rendered later).}
\label{fig:dag}
\end{figure}
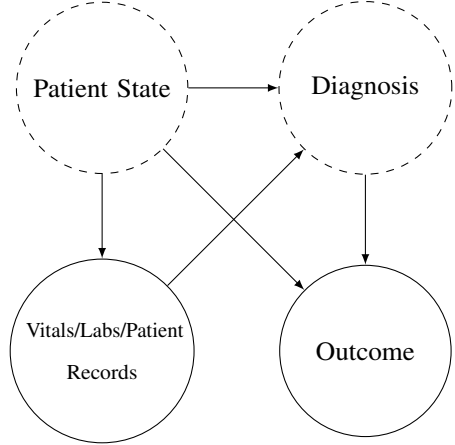

We begin with a simplified theory to help us grasp the main issues. Figure \ref{fig:dag} shows a simplified DAG of the data generating process. A patient has a true `state' at intake time which can be thought of as all the medically relevant characteristics of the patient. We do not observe this, rather we observe a proxy given by our labs and vitals. A medical team observes the labs and vitals as well as potentially other data which gives a diagnosis. The diagnosis (and subsequent treatment plan) combined with the patient state determine the outcome which we observe.

Here we see that the ACM is missing important information relevant to predicting the outcome. We now ask how this affects the reliability and usefulness of the model.

We refer to the vitals/labs/patient record features as $x$ and the diagnosis as $d$. For simplicity let the cardinality of $D$, the set of all diagnoses, be finite. We consider the case of binary outcome $y$ to make things simpler. Let $y=1$ be the `bad' outcome.

Given that the state is unobservable, the best we could attempt to predict the patient's outcome with is a model which learns the probability distribution $p(y=1 | x=x, d=d)$. 

The ACM learns instead the probability $p(y=1 \mid x)$. Note that this can be factorized as $$p(y=1 \mid x=x) = \sum_{d} p(y=1 \mid x=x, d=d) p(d \mid x=x).$$

In other words, the ACM is a weighted sum of diagnosis-specific probabilities weighted by the probability of $d$ given $x$. 

The factorization lays bare the main issue with the ACM. Consider the situation where two patients enter the hospital with the same $x$ but one has a standard infection and the other has an extremely antibiotic resistant version of the same. If we do not know which patient is which, medical prudence dictates that we treat both as high risk patients until we know otherwise.

Our key conceptual contribution is to instead consider the probability distribution implied by $$\hat{p}(y=1 \mid x=x, d=d) \hat{p}(d \mid x=x)$$ where $\hat{p}$ are trained models  For example, considering the risk of the patient as if they have the `worst' diagnosis that has some `reasonable' level of posterior probability.

We refer to this as a \textbf{diagnosis uncertain ACM} or DU-ACM.

\section{ACM vs. Diagnosis-Specific Models}
Let us ground the theoretical discussion above with an empirical analysis.

For all of our analyses we use the full MIMIC-IV dataset \citep{johnson2023mimic}. We restrict to only patients that appear in the dataset once.

For our patient features we get the first 24 hours of vital signs (blood pressure, heart rate, temperature, respiration rate, oxygen saturation,  and use the min, max, mean of these vitals. We also take the first 24 hours of lab results. We choose labs that appear in at least $85 \%$ of records (hematocrit, hemoglobin, platelets, wbc, aniongap, bicarbonate, blood urea nitrogen, calcium, chloride, creatinine, glucose, sodium, potassium, inr, pt, ptt) and take the min and max of these labs. We also use patient background features including comorbidities at the time of admission (prior myocardial infarction, congestive heart failure, peripheral vascular disease, cerebrovascular disease, dementia, chronic pulmonary disease, rheumatic disease, paraplegia, renal disease, malignat cancer, severe liver disease, metastatic solid tumor, and AIDs), age, ethnicity, and gender.

This gives us $30,938$ complete records, with 74 total features. We use in hospital mortality ($12.6$ percent of all patients) as our outcome variable.

In MIMIC each patient is associated with possibly multiple diagnoses. We take the primary diagnosis associated with each patient. 

\begin{figure}[h!]
    \centering
    \includegraphics[scale=.75]{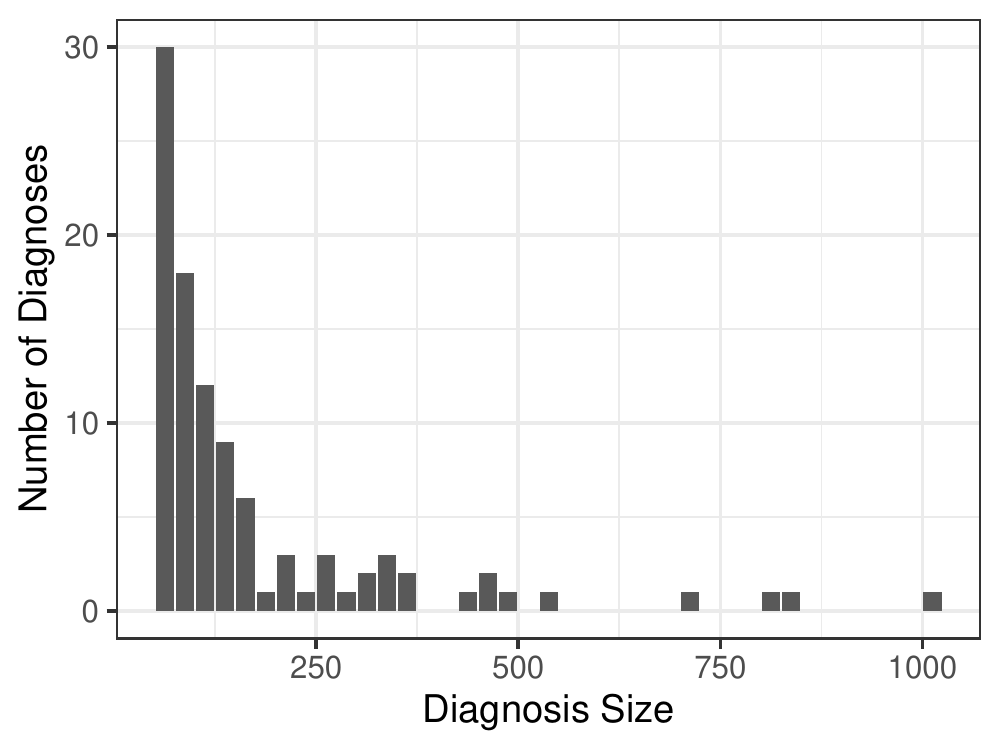}
    \caption{Most diagnoses have relatively few records in our dataset so diagnosis-specific models are often infeasible. When we study ACM vs. diagnosis-specific models we look at common diagnoses ($>200$ patients in the dataset). In the DU-ACM section we expand to all diagnoses with more than 50. }
    \label{fig:diag_size}
\end{figure}

There are 2807 distinct diagnoses with each having on average 11 patients. Figure \ref{fig:diag_size} shows the full distribution of patients with each diagnosis. We see that some diagnoses are quite rare but there is a substantial set of common ones. This distribution means that in general diagnosis-specific models may be infeasible for all but a small number of `head' diagnoses. 

Still, it is useful to compare an ACM to diagnosis specific modeling. We will consider the set of diagnoses with more than $200$ patients and a mortality rate greater than $.05$. These diagnoses are: acute kidney failure, acute myocardial infarction, acute respiratory failure, cerebral embolism, cerebral infarction, closed fracture of base of skull, infection and flammatory reaction, intracerebral hemorrhage, NSTEMI, non-traumatic intracerebral hemmorhage, pneumotitis, STEMI, secondary malignant neoplasm, sepsis (unknown organism), subarachnoid hemorrhage, subdural hemorrhage, subendocranial infarction, traumatic subdural hemorrhage, uspecified septicemia. We refer to this set as $D_{common}.$

\subsection{Transferrability of ACM Predictions Across Diagnoses}
We now compare the ACM to diagnosis-specific models in $D_{common}$. Our main contribution is agnostic to the model architecture used to fit it (e.g. logistic regression vs. neural network vs. random forest), however, we will still study this question here as it provides a useful view on the structure of our datasets and our problem.

We look at three potential model architectures. As a baseline we use a regularized logistic regression. 

We compare this to an generalized additive model (GAM \citet{hastie2017generalized}) which continues to be additive across features but is non-linear. Letting $x_j$ be an arbitrary element of the patient features a GAM models $$p(y=1 | x=x) = \sigma (\sum_{j} f_j (x_j)$$ where $\sigma$ is the standard sigmoid and $f_j(\cdot)$ are learned functions. 

There are many ways to fit GAMs, we focus on a simple, well studied one - EBM implemented via \textit{interpretml} \citep{nori2019interpretml}. An EBM is more accurate than a logistic regression since it does not assume linearity in features, but maintains an easy interpretability since the entire model can be summarized by the functions $f$ which can be plotted as $1-d$ risk curves which are simply added up.

Finally, we compare both of these to a full black box XGBoost model \citep{chen2015xgboost}. XGBoost is a more powerful model class (since it allows for various interactions between features unlike our GAM) but it loses interpretability.

As our evaluation we look at ROC-AUC in a $20 \%$ held out test set. 
For the regularized logistic regression we use cross-validation to set the regularization parameter using the package \textit{glmnet} \citep{hastie2014glmnet}. In the EBM we use the \textit{interpret} \citep{nori2019interpretml} package which uses bagging to control overfitting. We use 16 inner and 4 outer bags. For XGBoost we perform a cross-validated grid search over num trees, learning rate, and tree depth using the cv function in the \textit{XGBoost} R package.

\begin{figure}[h!]
    \centering
    \includegraphics[scale=.75]{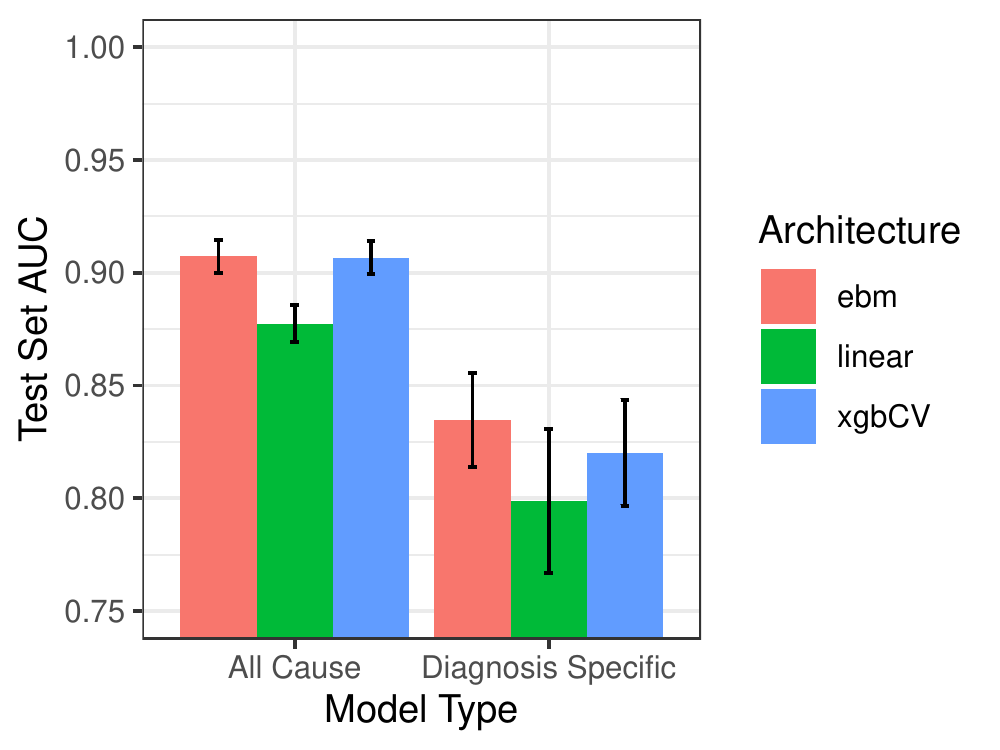}
    \caption{ACM vs. diagosis specific model AUCs on held-out test sets. Error bars on ACM reflect standard errors using AUC equivalence to the U statistic. Error bars on diagnosis-specific models are standard errors of the average across diagnoses.}
    \label{fig:model_comparo}
\end{figure}

Figure \ref{fig:model_comparo} shows that the ACM outperforms disease specific models on average across the full dataset. In addition, we see that XGBoost and EBM have similar performance while the logistic regression falls behind. For this reason we now turn to using only the EBM for model fitting.

We now discuss the ACM's performance in terms of bias and variance. Clearly, the ACM is trained on a much larger dataset so it has lower variance. We now investigate to what extent bias properties of the ACM can lead practitioners astray.

We begin by looking at whether the ACM can generalize across diagnoses well. To show this starkly, we consider a slightly different training setup which we refer to as out of diagnosis prediction. 

For each $d \in D_{common}$ we compare the predictive power of a model that is trained on all data \textit{except} $d$ to a model trained \text{only on $d$}. 

In the former case we use the entire set of patients with diagnosis $d$ for testing (since the model is trained on all patients with diagnosis not $d$. In the latter case we split the diagnosis data into $80 \%$ train and $20 \%$ test. 
Figure \ref{fig:diagvsac} shows our results. We see no cases where the diagnosis specific model outperforms the out of diagnosis model. This suggests there is high transferrability in information across diagnoses.

We perform one more experiment to understand generalization in the ACM. We look at all patients that have diagnoses not in $D_{common}.$

We consider all patients with $d \notin D_{common}$. We then train each of our disease-specific models for each $d \in D_{common}$ and look at the risks they assign to these held out patients. 

\begin{figure}[h!]
    \centering
    \includegraphics[scale=.75]{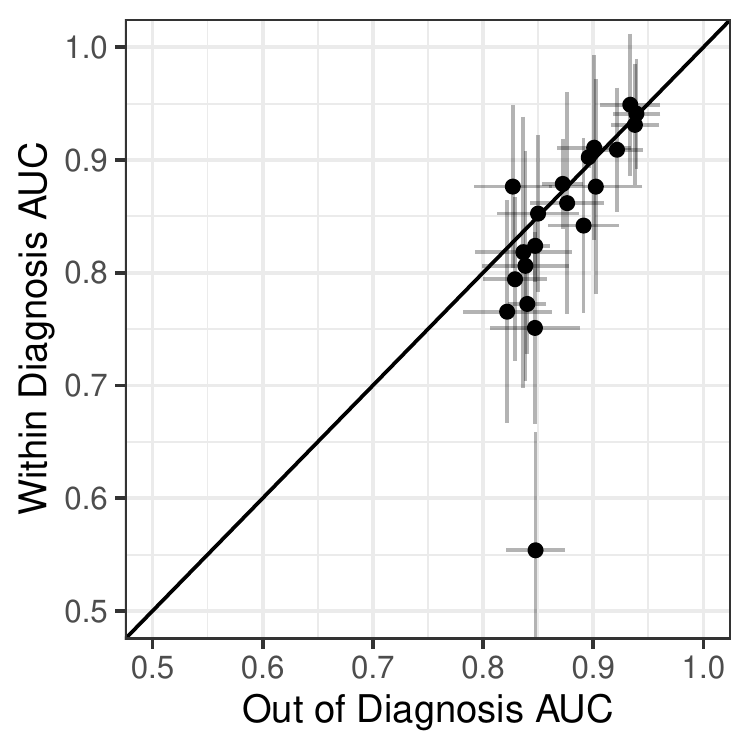}
    \includegraphics[scale=.75]{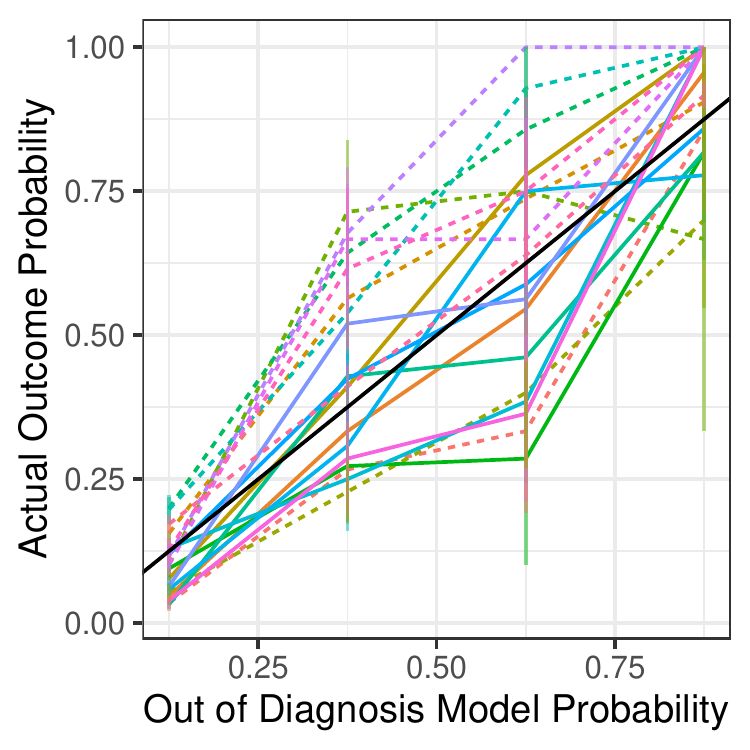} \\
    \includegraphics[scale=.5]{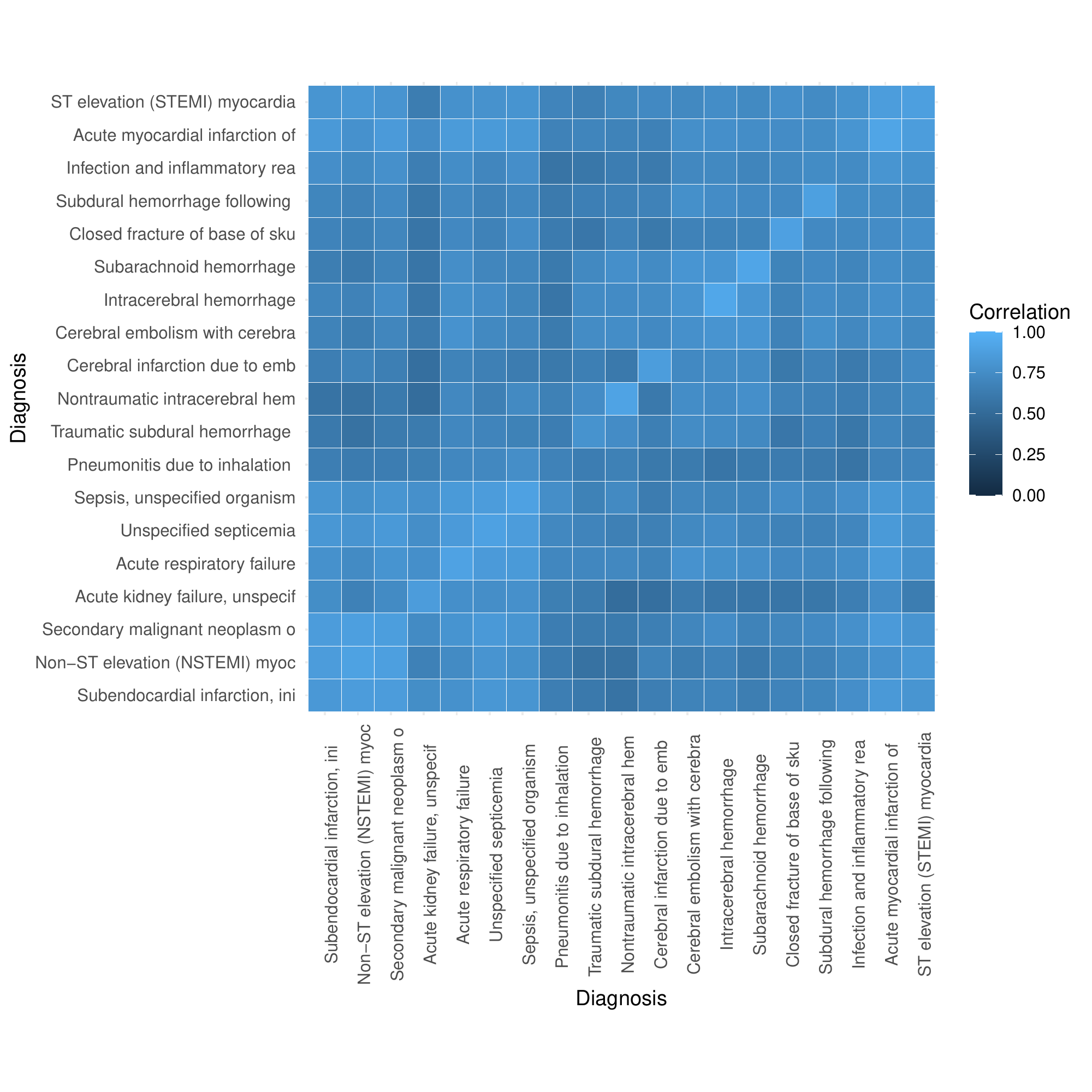}

    \caption{Panel 1: Each point represents a single diagnosis, a point below the 1-1 line means out-of-diagnosis AUC is higher than within-diagnosis AUC. Error bars use the equivalence of AUC to the U statistic. Panel 2: Calibration curves for out-of-diagnosis models. Dashed lines indicate that average calibration is not statistically distinguishable from $0$ at $p<.05$. Panel 3: Correlations between predictions of different within-diagnosis models on a held-out set. There is substantial similarity in what is learned across diagnosis-specific risk models as we see a $.85$ correlation between bootstrapped predictions (same diagnosis) and $.71$ correlation across models from different diagnoses.}
    \label{fig:ood}
\end{figure}

Figure \ref{fig:ood} plots the correlation between the predicted risks for each of the models. On the diagonal we plot the rank correlation in risks from the same model trained twice on a bootstrapped sample of its training data. We see that on average there is a $.85$ correlation between bootstrapped predictions and $.71$ correlation across models of different diagnoses. Thus, there is substantial agreement across diagnosis specific models.

However, while there is substantial cross-diagnosis generalization we see that it is not perfect. Panel 2 of Figure \ref{fig:ood} shows calibration curves for some diagnoses. We see that there appears to be a base-rate shift, for some diagnoses the entire risk curve is shifted up or down. These differences are not just due to random variation. \alex{discuss}, we see that even after adjusting for multiple comparisons via a  Benjamini-Hochberg procedure \citep{benjamini1995controlling} some diagnoses remain uniformly riskier (eg. intracerebral hemorrhage) or less risky (eg. infection and inflammatory reaction) than would be suggested by an out-of-diagnosis model.

\section{The DU-ACM}
This last section suggests unless $x$ perfectly separates patients by diagnosis, some predictions of an ACM will be too optimistic. We now turn to a way to exploit the generalization properties of the ACM without `averaging away' the risk of some patients. 
In essence, we seek to output the predictions for each patient but show the uncertainty that remains just due to diagnosis. Algorithm \ref{alg:duacm} shows our basic idea. 

\begin{algorithm}
\caption{Diagnosis Uncertain All Cause Model Algorithm}\label{alg:duacm}
\begin{algorithmic}
\Require Training time data with $(x_i, d_i)$ for each patient $i$
\Require Inference time data with just $(x_j)$ for each patient $j$

\While{Training}
\State Train an outcome model $f(x, d)$ which solves $\min \text{Loss}(y_i, f(x_i, d_i))$
\State Train a diagnosis model $g(x)$ which solves $\min \text{Loss}(d_i, f(x_i))$
\EndWhile

\While{Inference for patient $j$}
\State $y_{sampled} = [ ]$
\For{num samples}
\State Use diagnosis model to sample $\hat{d}_j$ from $g(x_j)$
\State Append $f(x_j, \hat{d}_j)$ to $y_{sampled}$
\EndFor
\State Use $y_{sampled}$ as estimate of distribution of possible outcomes
\State Return distribution of $y$ as well as explanations (e.g. list of model predicted risky diagnoses)
\EndWhile
\end{algorithmic}
\end{algorithm}

Note that the Algorithm \ref{alg:duacm} is agnostic to the model class used to fit $f$ and $g$. So far we have seen that if in a diagnosis $d$ a patient with profile $(x, d)$ is more risky than $(x', d)$ then given another diagnosis $d'$ we still have that $(x, d')$ is more risky than $(x', d')$. 

This makes intuitive medical sense - for example, having low blood oxygen saturation is worse than good blood oxygen saturation no matter if one is in the hospital for a blunt force trauma or a heart attack.

If we assume that the risk ordering is approximately equal across diagnoses, it means we can keep our additive framework to approximate the full probability as $$p(y=1 \mid x, d) = \sigma(\sum_{j} f_j (x_j) + \beta(d))$$ where $\beta(d)$ is an offset (or base rate) for each diagnosis. Such a model is exactly in the framework of the EBM. So, we will fit $f$ using EBM. For diagnoses, we use all diagnoses with more than $50$ patients (94 total diagnoses).

To fit $g$ we use a neural network with $2$ hidden layers with $64$ hidden units. We found tree-based models to do poorly in the many class $d$ regime both in terms of convergence speed and overall classification accuracy. We split the training set into a train and a validation set to choose hyperparameters (learning rate, weight decay) then train a model on the combined set and predict to a completely held out test set. Our NN achieves a one-vs-all held out AUC of $.86$ in the test set.

We now consider using the DU-ACM at inference time on a held out test set. For a patient $j$ we sample $150$ diagnoses conditional on $x_j$ using $g(x_j)$. Plugging these sample into $f(x_j, d_j)$ gives us a distribution of probabilities. We then take the mean (which corresponds to approximating the ACM) - we refer to this as $\hat{y}$ and the $90^{th}$ percentile $\hat{y}^{90}$ which corresponds to a pessimistic prediction.

\begin{figure}[h!]
    \centering
    \includegraphics[scale=.75]{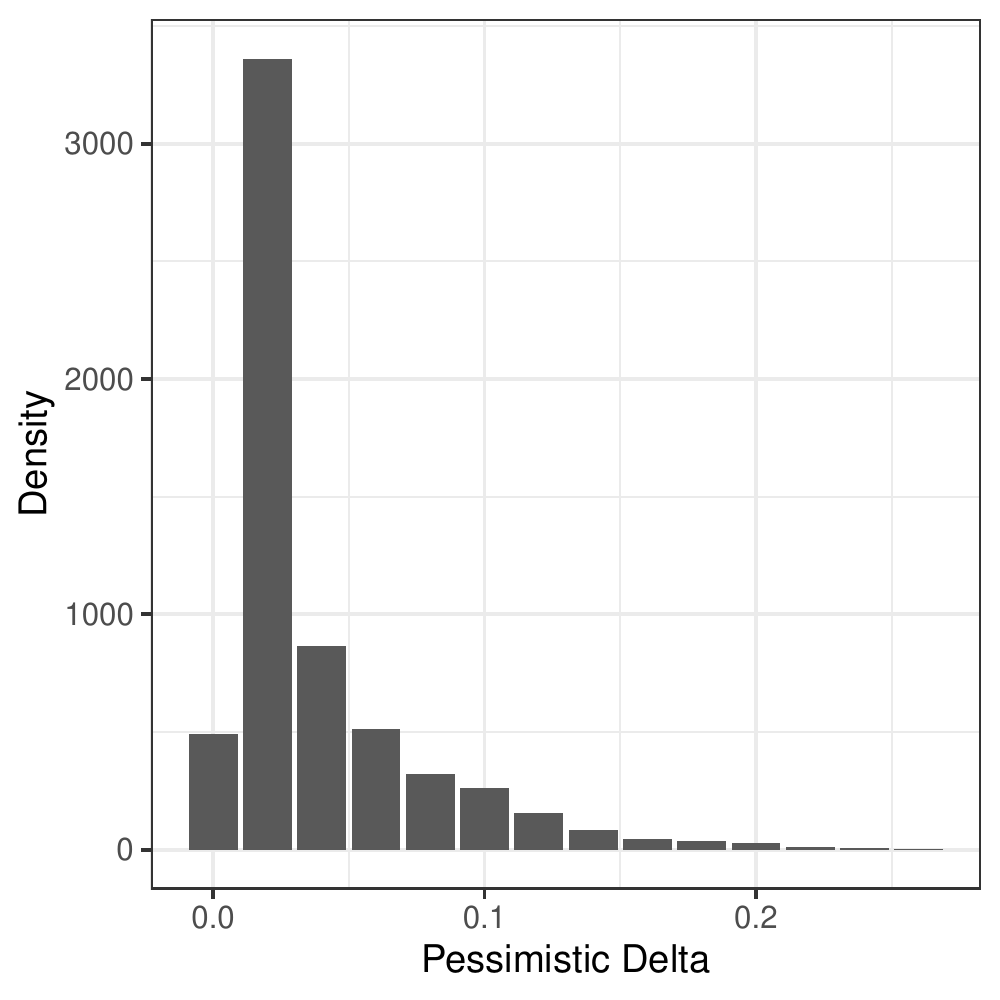}
    \includegraphics[scale=.75]{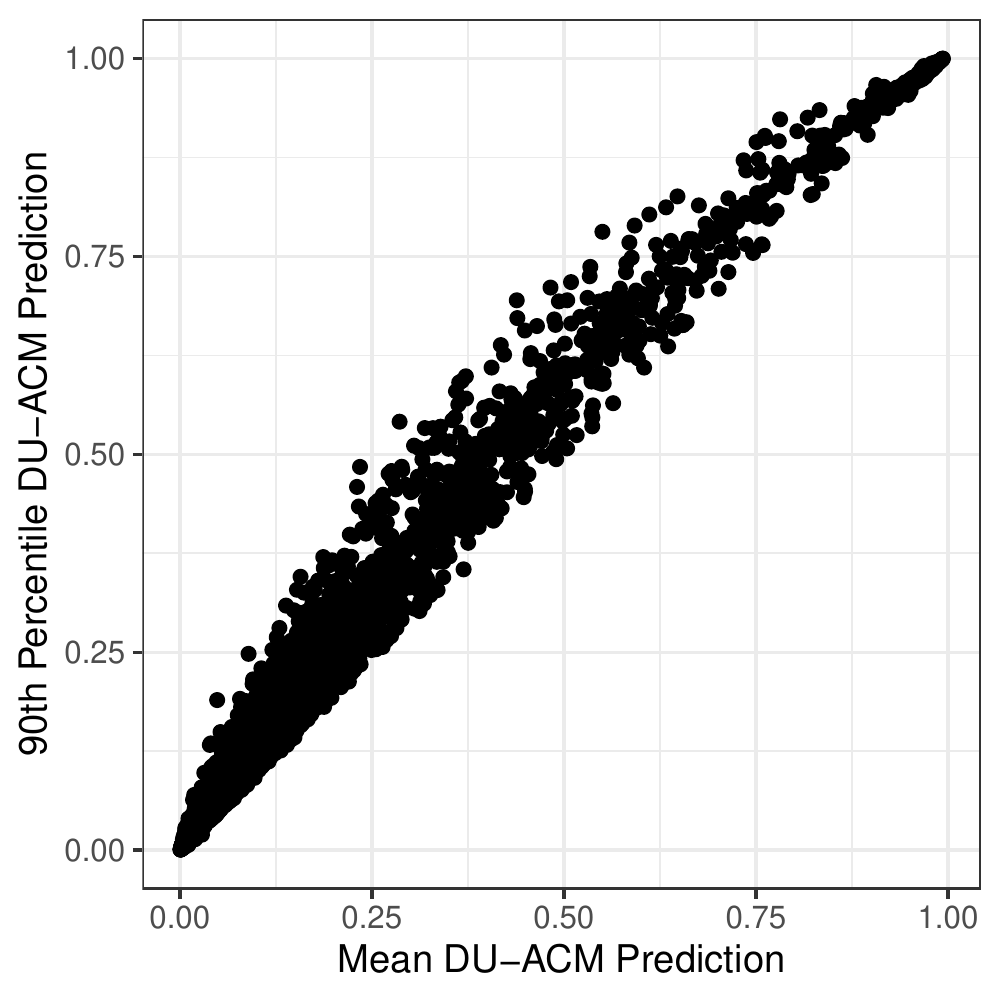}
    \caption{Histogram of Q90 - mean DU-ACM predictions show that there is lack of uncertainty about most patients but substantial diagnosis-created uncertainty for some. Right panel shows, as we might expect, that uncertain patients are situated near the middle of the risk curve.}
    \label{fig:du_acm}
\end{figure}

In Figure \ref{fig:du_acm} we plot what DU-ACM predictions look like. The first panel gives us a distribution of $\hat{y}^{90} - \hat{y}$ which we refer to as the `pessimistic delta' in the test set. 

The second panel shows the relationship of the gap with mean risk. A vast majority of patients have a very small gap, but there is a set of patients that may be classified as `safe' patients in the average case but risky in the worst case.

\begin{figure}[h!]
    \centering
    \includegraphics[scale=.8]{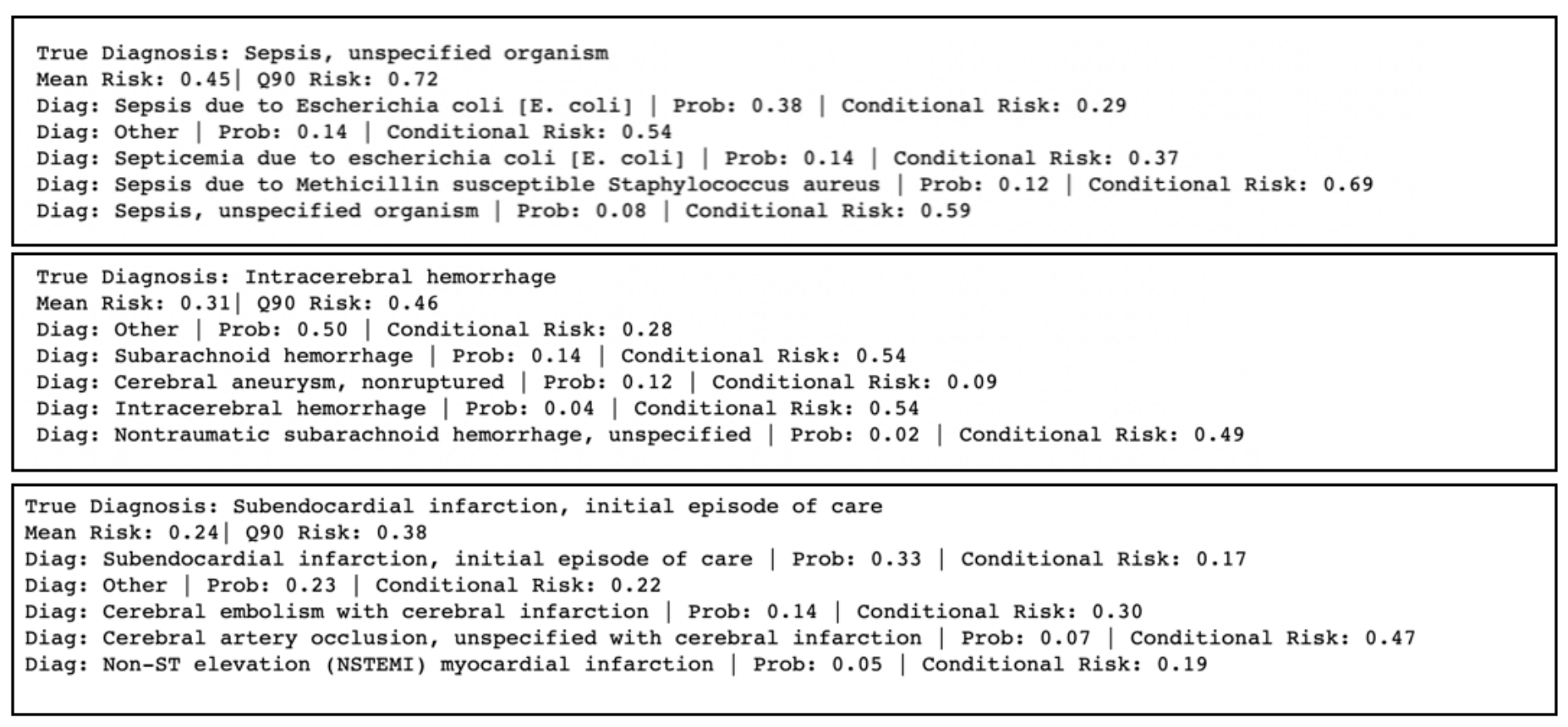}
    \caption{An example of a UI that could be used in a production version of a DU-ACM with real test set examples. We show both the mean expected risk as well as a `pessimistic' Q90 risk. We also show an explanation for the output. Looking at the first example the diagnosis model guesses that the patient has some form of infection, if the infection is E.Coli Sepsis then the risk to the patient is relatively lower compared to a MRSA Sepsis.}
    \label{fig:du_acm_examples}
\end{figure}

We give some examples of patients with high pessimistic deltas in Figure \ref{fig:du_acm_examples}. This serves as an example of a UI that could be used in a production version of a DU-ACM. In addition to risk assessment the DU-ACM outputs some explanation. Looking at the first example the diagnosis model guesses that the patient has some form of infection but is unsure which one, if the infection is E.Coli Sepsis then the risk to the patient is low, however, if the infection is due to MRSA then the patient's risk is quite high.

This gives us another way of using the DU-ACM in practice. If practitioners can rule out certain diagnoses, we can update the DU-ACM predictions live by simply taking the probability model and setting some outputs to $0$.

\section{Conclusion}
We have shown that all-cause risk predictors perform well on average but can mask the risk of patients with high risk but rare diagnoses. To deal with this issue we have introduced the DU-ACM which tries to account for uncertainty in risk estimates coming from lack of knowledge about a patient's eventual diagnosis.

\newpage

\bibliographystyle{unsrtnat}
%\bibliography{references}
\bibliography{paper.bbl}  %%% Uncomment this line and comment out the ``thebibliography'' section below to use the external .bib file (using bibtex) .

\end{document}